\documentclass[conference]{IEEEtran}
\IEEEoverridecommandlockouts
\usepackage{cite}
\usepackage{amsmath,amssymb,amsfonts}
\usepackage{algorithmic}
\usepackage{graphicx}
\usepackage{textcomp}
\usepackage{xcolor}
\usepackage{todonotes}
\usepackage{framed}
\usepackage{float}
\usepackage[export]{adjustbox}
\usepackage{caption}
\usepackage{subcaption}

\usepackage[colorlinks=true, linkcolor=black,urlcolor=black, citecolor=black]{hyperref}
\def\BibTeX{{\rm B\kern-.05em{\sc i\kern-.025em b}\kern-.08em
    T\kern-.1667em\lower.7ex\hbox{E}\kern-.125emX}}
\begin{document}

\title{Jewelry Recognition via Encoder-Decoder Models\\

}

\newcommand{\newlineauthors}{%
  \end{@IEEEauthorhalign}\hfill\mbox{}\par
  \mbox{}\hfill\begin{@IEEEauthorhalign}
}
\author{
\IEEEauthorblockN{José M. Alcalde-Llergo}
\IEEEauthorblockA{
\textit{Dept. of Economics, Engineering,} \\
\textit{Society and Business Organization (DEIM)}\\
\textit{University of Tuscia }\\
Viterbo, Italy \\
jose.alcalde@unitus.it}
\and
\IEEEauthorblockN{Enrique Yeguas-Bolívar}
\IEEEauthorblockA{\textit {Computing and Numerical Analysis} \\
\textit{University of Córdoba}\\
Córdoba, Spain \\
eyeguas@uco.es} 
\and 
\IEEEauthorblockN{Andrea Zingoni}
\IEEEauthorblockA{
\textit{Dept. of Economics, Engineering,} \\
\textit{Society and Business Organization (DEIM)}\\
\textit{University of Tuscia }\\
Viterbo, Italy \\
andrea.zingoni@unitus.it}
\and
\IEEEauthorblockN{ \hspace{0.5cm}}
\IEEEauthorblockA{}
\and
\IEEEauthorblockN{Alejandro Fuerte-Jurado}
\IEEEauthorblockA{\textit {Backend Developer} \\
\textit{GAC Travel}\\
Córdoba, Spain \\
alejfuejur@gmail.com}

}

\maketitle

\begin{abstract}
Jewelry recognition is a complex task due to the different styles and designs of accessories. Precise descriptions of the various accessories is something that today can only be achieved by experts in the field of jewelry. In this work, we propose an approach for jewelry recognition using computer vision techniques and image captioning, trying to simulate this expert human behavior of analyzing accessories. The proposed methodology consist on using different image captioning models to detect the jewels from an image and generate a natural language description of the accessory. Then, this description is also utilized to classify the accessories at different levels of detail. The generated caption includes details such as the type of jewel, color, material, and design. To demonstrate the effectiveness of the proposed method in accurately recognizing different types of jewels, a dataset consisting of images of accessories belonging to jewelry stores in Córdoba (Spain) has been created. After testing the different image captioning architectures designed, the final model achieves a captioning accuracy of 95\%. The proposed methodology has the potential to be used in various applications such as jewelry e-commerce, inventory management or automatic jewels recognition to analyze people's tastes and social status.
\end{abstract}


\section{Introduction}
Image captioning is a challenging task in the field of computer vision and natural language processing, which involves generating a textual description for a given input image \cite{show_and_tell}. The goal of image captioning is to enable machines to understand visual content in a similar way as humans and generate captions that accurately describe the content of the image. This task has gained significant attention in recent years due to its potential applications in various domains, such as assistive technology for visually impaired individuals, automatic image tagging and retrieval, or social media content analysis.

In the last years, deep learning based approaches have emerged as a promising solution to the problem of image captioning, leveraging the power of deep neural networks to learn complex mappings between images and text \cite{survey}. In order to perform this task an encoder-decoder structure is commonly used. Specifically, convolutional neural networks (CNNs) are typically the encoders due to their ability to process the input image and extract a set of high-level feature representations. On the other hand, recurrent neural networks (RNNs) are commonly used as decoder to generate a sequence of words to describe the image defined by the features extracted by the encoder. For this reason, these two techniques working together have significantly improved the performance of image captioning systems, achieving an almost human-level performance on some benchmark datasets \cite{encoder-decoder}. Nevertheless, there are still many challenges and open problems that need to be addressed to further improve the accuracy and robustness of image captioning systems.

An artificial intelligence tool capable of automatically identifying and describing jewelry can simulate human intuitive behavior, leading to an increased connection between individuals from diverse cultures and customs:

\begin{itemize}
    \item Overcoming language and cultural barriers: by providing accurate and detailed descriptions of jewelry in various languages, an AI tool for jewelry recognition can overcome language and cultural barriers. This enables people to understand and appreciate the characteristics and cultural value of jewelry from other cultures.
    \item Encouraging curiosity and cultural exchange: by offering information about the history, symbolism, and craftsmanship of jewelry, a jewelry recognition AI tool can spark curiosity and foster deeper cultural exchange. This can lead to meaningful conversations and dialogues between individuals of different backgrounds.
    \item Promoting appreciation and respect for diversity: understanding and appreciating jewelry from different cultures can foster greater appreciation and respect for cultural diversity. This can promote tolerance, inclusivity, and harmony among individuals and communities with different traditions and customs.
    \item Facilitating cultural commerce and tourism: automatic jewelry recognition can have practical applications in cultural commerce and tourism. By providing accurate information about the authenticity and characteristics of jewelry, the tool can assist buyers and collectors in making informed decisions and engaging in fair trade of cultural jewelry. Moreover, it can attract tourists interested in exploring and acquiring traditional jewelry during their travels.
\end{itemize}

In conclusion, an AI tool for jewelry recognition has the potential to promote understanding, cultural exchange, diversity appreciation, and fair trade, thereby strengthening relationships between individuals from different cultures.

In this work, the performance of different image captioning encoder-decoder structures will be compared on a jewels images dataset generated by using images of different accessories obtained from two well-known online jewelry stores in the city of Córdoba (Spain). Córdoba is renowned for its artisanal craftsmanship and expertise in creating high-quality jewelry, particularly in silver. The city hosts significant jewelry fairs and events, showcasing unique designs and attracting both domestic and international buyers. 

 This task is complex and remarkably captivating, given the limited application of image captioning within this particular field of study thus far. The final objective of the project is to generate detailed captions about the accessories that appear in the input images, indicating characteristics such as their type (earrings, necklace, bracelet, etc.), material, color and the jewelry they contain. The images used include people wearing the accessories, so this system would also be used to automatically generate descriptions of accessories worn by a particular person in images from fashion magazines, jewelry catalogs, jewelry trade shows, etc. In addition, the methodology also allows to make simpler descriptions indicating only the type of the accessory in order to perform a jewelry multilabel classification task.

\section{Related work}
Within this section, we present a comprehensive examination of prior works that have served as influential sources and points of reference for the present study. We have identified two primary categories of research that are pertinent to our investigation: image captioning and the utilization of artificial vision techniques for jewelry analysis. Additionally, we have conducted a study specifically focusing on the application of image captioning techniques in the domain of jewelry; however, no previous research has been identified that specifically addresses this precise subject matter.

\subsection{Image captioning}
Describing the content of an image is a task that some years ago could only be done by humans. However, advances in computer vision have allowed to create new systems which are able to reproduce this human behavior obtaining pretty good performance.

The initial steps taken in this field were documented in works such as \cite{show_and_tell} in which the authors developed a probabilistic framework for the purpose of generating descriptions from images. To achieve this objective, they designed an encoder-decoder structure that employed a CNN as the encoder and a RNN as the decoder. This particular architecture has served as the foundation for image captioning, whereby the input image is processed through the CNN to encode and extract its features, which are then embedded into a fixed-length vector. Subsequently, the final descriptions are generated by appending the RNN to the last hidden layer of the CNN. Specifically, a Long Short-Term Memory (LSTM) network was selected as the RNN due to its efficacy in translating and generating novel sentences.


Similarly, numerous works have been developed subsequently to the aforementioned study, which have integrated novel mechanisms to enhance its outcome. Notably, one of the most interesting proposals was presented in \cite{show_attend_and_tell}, which incorporated the concept of attention into the previously utilized encoder-decoder architecture. In this instance, diverse words generated by the LSTM were focused on different sections of the input image. The results illustrated that their model generated highly precise words that were directly correlated to the areas of the image that the LSTM "paid attention" to at each point in time. Consequently, a comprehensive and precise description of the image was achieved in most cases.

\subsection{Computer vision for jewels}
Jewelry has been an essential part of human culture for thousands of years, and the ability to recognize different types of accessories is valuable in various fields, including security, e-commerce and analyze social tastes. Computer vision models have come a long way in recent years, with significant advances in object recognition and detection. There are many models available that can accurately identify different types of objects, such as animals, vehicles, and even accessories. An example of these models is DeepBE, presented in \cite{reco_jewels_face}, which is able to classify the accessories that a person wears over the shoulders (necklaces, earrings, glasses, etc.). However, current methods for accessory recognition as the one mentioned previously remain somewhat limited compared to a human expert. While these models can accurately identify the presence of jewelry and other accessories in an image, they are unable to provide detailed information regarding the type of jewelry or its quality. As a result, there is a need for more specialized models that can accurately and comprehensively recognize and classify jewelry and other accessories based on their unique characteristics.

On the other hand, some studies have been dedicated to the development of specialized models for assessing the quality of specific types of jewelry. For instance, the work presented in \cite{pearls_quality} focuses on evaluating the quality of pearls based on images, utilizing computer vision techniques to identify and analyze the unique visual characteristics of these objects. The proposed approach entails tracing light rays and analyzing highlight patterns in order to estimate the level of specularity present in the pearl given as input image. This information is then utilized to determine the equivalent index of appearance, which serves as a metric for evaluating the quality of the object. By utilizing these techniques, the method was able to comprehensively assess the visual appearance of the object, providing valuable insights into its quality and potential industrial applications.

\section{Experimental design}

To achieve the objective of describing accessories and jewellery present in the images, we will begin by providing an explanation of how the datasets utilized to train, validate, and test the models were created. Subsequently, the chosen encoder-decoder architectures used in combination to perform the image captioning task will be presented.

\subsection{Datasets}
The datasets used during the experiments have been created specifically for this work, by extraction, preparation, and merging of images from two online jewelry stores in Córdoba. The stores are Baquerizo Joyeros \footnote{Baquerizo, “Baquerizo joyeros 1945.” \url{https://baquerizojoyeros.com/}.} and Doñasol \footnote{DoñaSol, “Doñasol joyas.” \url{https://www.doñasol.com/}}. However, as with most computer vision problems, we did not have enough images to train sufficiently robust models to obtain good results. This is why we decided to pursue several data augmentation techniques, listed below:

\begin{itemize}
    \item 90º rotations.
    \item Image width shift by 30\%.
    \item Image length shift by 30\%.
    \item Cuts the image by 15\%.
    \item Image enlargement or reduction by 5\%.
    \item Slight color changes.
    \item Horizontal and vertical flips.
    \item Brightness range by 80\%.
\end{itemize}

Finally, after the data augmentation, the merging process resulted in a comprehensive and unified database featuring a total of 2687 accessory images. For experimentation, this dataset has been divided as follows: 75\% for training set, 15\% for validation set and 10\% for testing set. Some image samples from the dataset are shown in Figure \ref{fig:joyas}, with the relative captions.

\begin{figure*}
        \centering
        \begin{subfigure}[b]{0.3\textwidth}
             \centering
             \includegraphics[width=\textwidth]{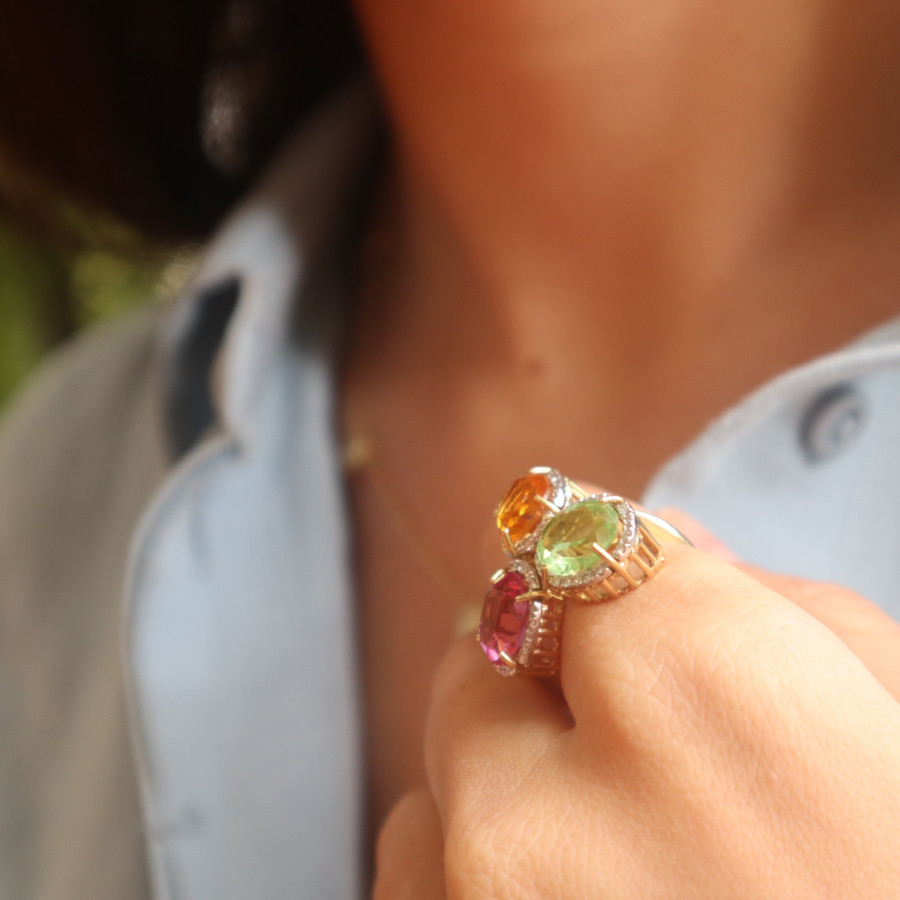}
             \caption{Skye yellow gold and oval stones ring.}
             \label{subfig:ring}
        \end{subfigure}
        \begin{subfigure}[b]{0.3\textwidth}
             \centering
             \includegraphics[width=\textwidth]{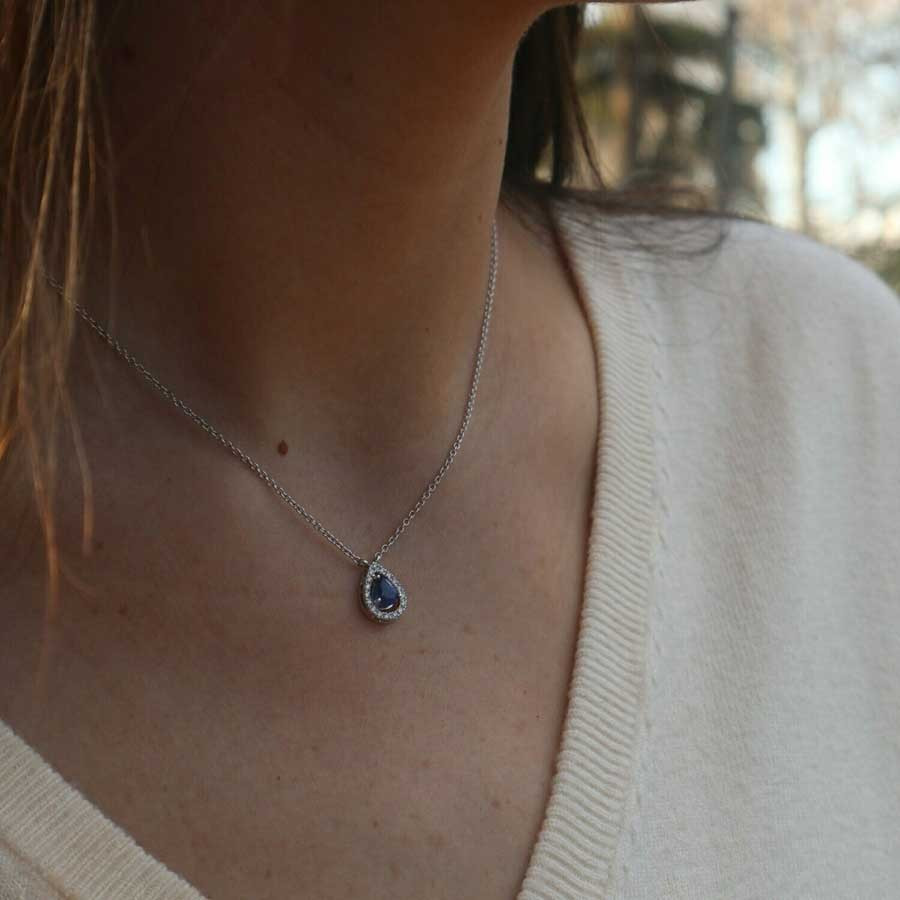}
             \caption{Arayat emerald necklace.}
             \label{subfig:necklace}
        \end{subfigure}
        \begin{subfigure}[b]{0.3\textwidth}
             \centering
             \includegraphics[width=\textwidth]{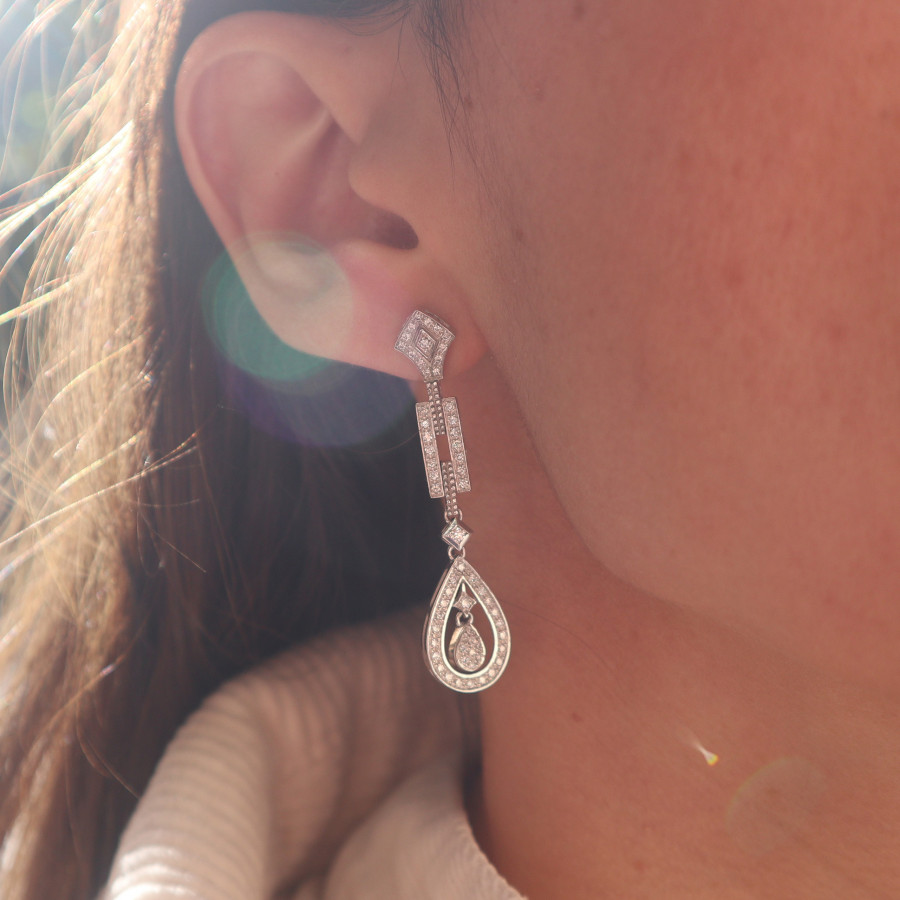}
             \caption{Taurus white gold, diamonds earring.}
             \label{subfig:earrings}
        \end{subfigure}
        \caption{Training images and captions from the final dataset.}
        \label{fig:joyas}
    \end{figure*}

\subsection{CNN and RNN architectures}

Thanks to the existence of pre-trained convolutional networks with vast amounts of data, we can utilize them in order to achieve improved outcomes. Consequently, it is not necessary to develop a convolutional network from scratch. This technique of network reuse is referred to as Transfer Learning \cite{transferLearning}.

During the experimentation step different CNNs and RNNs architectures have been combined to find the best encoder-decoder structure to describe the accessory images of the dataset. Regarding CNN, three popular architectures have been considered:
\begin{itemize}
    \item \textbf{VGG-16} \cite{vgg16}. It was one of the first CNNs to appear and one of the most popular ones. It is a fairly simple architecture, using only blocks composed of an incremental number of convolutional layers with filters of size 3×3. 
    \item \textbf{InceptionV3} \cite{inceptionv3}. It was developed by Google and achieved state-of-the-art performance on a number of computer vision tasks, including image classification and object detection. In addition, this model also obtains pretty good result in quite difficult medical image analysis tasks, as can be seen in works such as \cite{inc3_alzheimer}, \cite{inc3_covid}, \cite{inc3_diseases}.
    \item \textbf{MobileNet} \cite{mobilenet}. This architecture was developed and tuned to be used in mobile applications. The reason for choosing it is to check if it would be feasible to launch the type of structure we are developing in smartphones, as it could be very interesting for our problem.
     
\end{itemize}

On the other hand, RNNs have been widely used for sequence modeling and predictions. However, one of their major limitations is their susceptibility to short-term memory loss, which can impact the network's ability to capture long-term dependencies in the data. To address this limitation, more advanced models such as LSTM \cite{LSTM} and Gated Recurrent Unit (GRU) \cite{gru} were introduced. These models incorporate mechanisms called gates, which are neural networks designed to control the flow of information through the sequence chain. By regulating the flow of information, these models can mitigate the effects of short-term memory loss and improve the network's performance on long sequences. That is the reason why these two RNN architectures have been chosen as decoders for the different proposed architectures in this work. 

\section{Experiments}
Other parameters have also been considered apart from the CNN architecture and the type of RNN. These parameters have been:
the number of neurons, number of epochs, optimization and batch size. 

\begin{itemize}
    \item Number of neurons of the RNN. Considered values: 64, 128, 256, 512 and 1024.
    \item Batch size. Considered values: 4, 8, 32, 128 and 512.
    \item Use of an optimizer. Considered optimizers: Adam, Adagrad, Adadelta and RMSProp.
    \item Learning rate. Considered values: 0.0001, 0.001, 0.01, 0.1.
\end{itemize}

Moreover, in order to avoid overfitting, an early stopping have been applied if the results for the validation set do not improve in a determined number of epochs. 

Different experiments have been designed to perform a robust comparison of the different parameter configurations to generate the models. After these experiments the best model configuration to obtain captions from the jewelry dataset has been computed.

In terms of measuring the quality of captions, given the intricate nature of image caption evaluation, determining appropriate assessment criteria poses significant challenges. Numerous metrics currently exist to measure the quality of language and semantic accuracy of captions. Some of the most commonly used evaluation metrics are METEOR, BLEU or ROUGE \cite{metrics}. Nevertheless, considering the constraints posed by the descriptions in our dataset, we have opted to consider those generated captions that match the original ones as correct. For future implementations, we intend to enhance the dataset and employ the aforementioned metrics to comprehensively assess the performance of our model.


\section{Results and discussion}
In this section some of the best results obtained are shown. As a first and simpler task, a classification of the kind of accessory have been performed, distinguishing among four classes: necklaces, rings, earrings and bracelets. After analyzing the best configuration for each of the encoder-decoder considered structures, the best results in terms of test CCR have been achieved by using VGG-16 as CNN and GRU as RNN, as shown in Table \ref{tab:class}. This table also shows other metrics obtained that have been considered of interest, such as validation CCR and Loss. Once the best configuration has been selected, a more exhaustive analysis was performed by using more metrics; as precision, recall and F1-Score; to study how good the model was at classifying each class of jewels individually. Table \ref{tab:ind_class} shows that the model is quite good at the classification task, scoring above 94\% for all metrics and accessory types, except for bracelets, which have been more difficult to detect.

\begin{table}[h]
\centering
\caption{Best model configuration for classification}
\label{tab:class}
\begin{tabular}{c|c|c|c|c|c}
CNN             & RNN          & Neurons         & Val. CCR        & Val. Loss       & Test CCR        \\ \hline
InceptionV3     & LSTM         & 512           & 0.9885          & 0.0334          & 0.9044          \\
VGG-16          & LSTM         & 256           & 0.9846          & 0.0562          & 0.8824          \\
MobileNet       & LSTM         & 64            & 0.9568          & 0.2249          & 0.9118          \\
InceptionV3     & GRU          & 1024          & 0.9885          & 0.0407          & 0.9044          \\
\textbf{VGG-16} & \textbf{GRU} & \textbf{1024} & \textbf{0.9789} & \textbf{0.1066} & \textbf{0.9118} \\
MobileNet       & GRU          & 64            & 0.9539          & 0.2827          & 0.8824         
\end{tabular}
\end{table}

\begin{table}[h]
\centering
\caption{Individual class classification}
\label{tab:ind_class}
\begin{tabular}{c|c|c|c}
Accessory & Precision & Recall & F1-Score \\ \hline
Necklaces & 0.9459    & 1      & 0.9722   \\
Rings     & 0.9762    & 0.9762 & 0.9762   \\
Earrings   & 0.9778    & 0.9778 & 0.9778   \\
Bracelets & 0.9167    & 0.7857 & 0.8462  
\end{tabular}
\end{table}

It is also important to point out that an optimization of the different parameters considered has been carried out, obtaining the best results with the following configuration:

\begin{itemize}
    \item Number of neurons: 1024.
    \item Batch size: 8.
    \item Optimizer: Adam.
    \item Learning rate: 0.001.
\end{itemize}

Concerning the main task of generating complete captions from accessories images, the best results were also obtained by using the VGG-16 and GRU combination. MobileNet was excluded from this study due to its inadequate performance in handling complex problems. Table \ref{tab:img_cap_best} shows the comparison between all encoder-decoder combination considered for this task. 

\begin{table}[h]
\centering
\caption{Image captioning best models}
\label{tab:img_cap_best}
\begin{tabular}{c|c|c|c|c|c}
CNN             & RNN          & Neurons        & Val. CCR        & Val. Loss       & Test CCR        \\ \hline
InceptionV3     & LSTM         & 1024         & 0.9829          & 0.0682          & 0.8857          \\
VGG-16          & LSTM         & 1024         & 0.9727          & 0.0890          & 0.9143          \\
InceptionV3     & GRU          & 256          & 0.9659          & 0.0968          & 0.8714          \\
\textbf{VGG-16} & \textbf{GRU} & \textbf{256} & \textbf{0.9829} & \textbf{0.0716} & \textbf{0.9571}
\end{tabular}
\end{table}

In this experimentation an optimization of the different parameters considered has also been performed, obtaining the best results with the following configuration:

\begin{itemize}
    \item Number of neurons: 256.
    \item Batch size: 16.
    \item Optimizer: Adam.
    \item Learning rate: 0.001.
\end{itemize}

In addition, in order to check whether early stopping was causing the model not to achieve a better accuracy, experimentation was run for 200 epochs by disabling it. Figure \ref{fig:ES_check} shows how the accuracy and loss metrics are not improved from around the epoch 40. This indicates that the use of early stopping has been a profitable technique to avoid overfitting and to reduce the computational time.

\begin{figure}[h]
        \centering
        \begin{subfigure}[b]{0.47\textwidth}
             \centering
             \includegraphics[width=\textwidth]{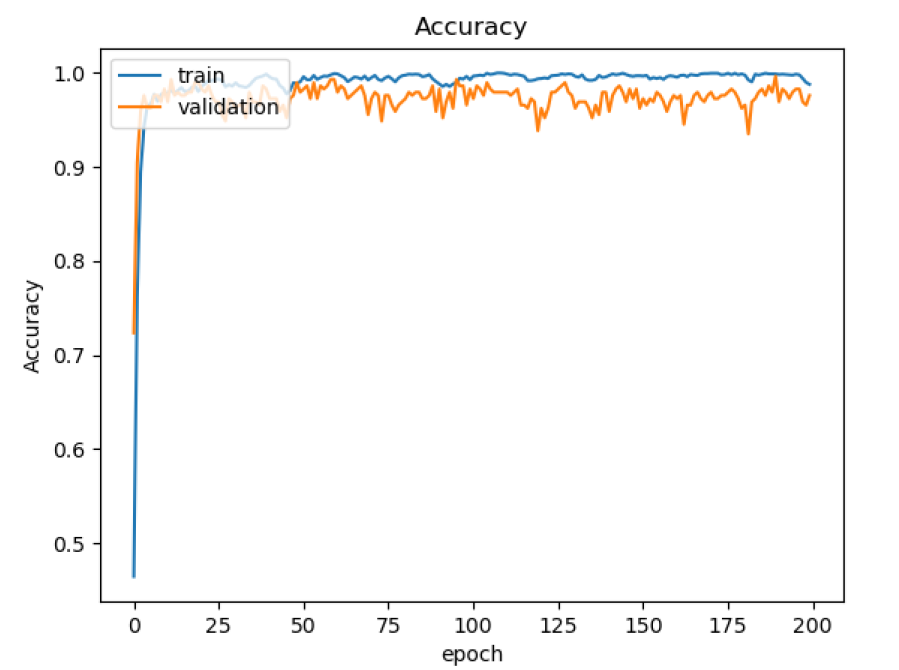}
             \caption{Train accuracy vs. validation accuracy considering only the first 200 epochs.}
             \label{subfig:ES_acc}
        \end{subfigure}
        \begin{subfigure}[b]{0.47\textwidth}
             \centering
             \includegraphics[width=\textwidth]{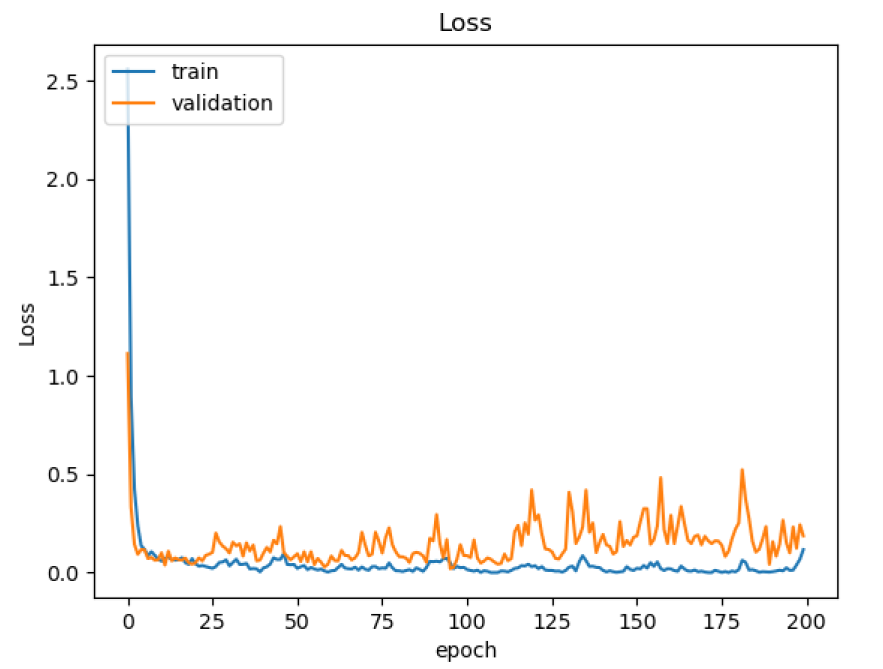}
             \caption{Train loss vs. validation loss in considering only the first 200 epochs.}
             \label{subfig:ES_loss}
        \end{subfigure}
        \caption{Validating the use of Early stopping by considering only the first training epochs.}
        \label{fig:ES_check}
    \end{figure}

Furthermore, the evaluation of our proposed approach demonstrated that the best-performing model achieved perfect captioning accuracy for all types of rings, necklaces, and bracelets. However, the classification accuracy for earrings was lower, with 87.5\% of the test samples being classified correctly. This accuracy has been calculated by comparing whether the generated captions exactly matched the real descriptions of the accessories in the image. In addition, the misclassifications observed were typically attributed to accessories with similar shapes and materials, as illustrated in Figure \ref{fig:mal_1}, or cases where the same type of accessory could be composed by varying materials or jewels, as illustrated in Figure \ref{fig:mal_2}.

\begin{figure}[h]
        \centering
        \begin{subfigure}[b]{0.23\textwidth}
             \centering
             \includegraphics[width=\textwidth]{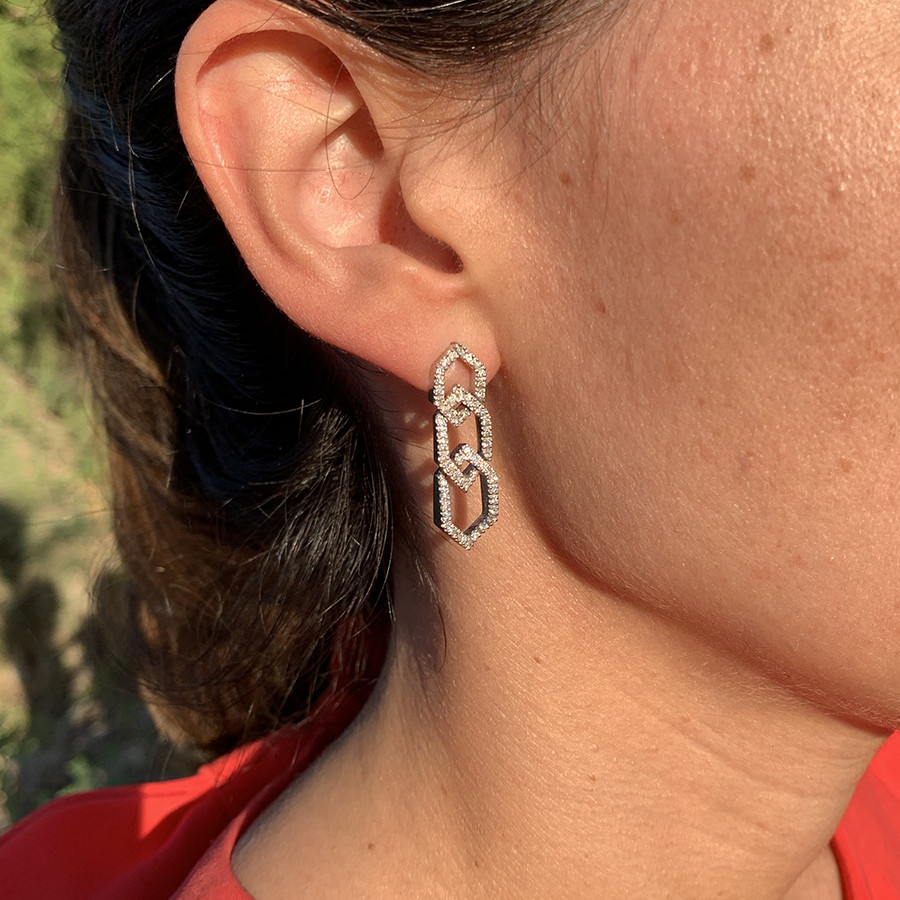}
             \caption{Badly descripted accessory ``Colette diamond earring''.}
             \label{subfig:bad1}
        \end{subfigure}
        \begin{subfigure}[b]{0.23\textwidth}
             \centering
             \includegraphics[width=\textwidth]{img/data/pendientes-taurus_2.jpg}
             \caption{Descripted as ``Taurus white gold, diamonds earring''.}
             \label{subfig:bad2}
        \end{subfigure}
        \caption{Accessories badly described by the best model. Similar shape and material.}
        \label{fig:mal_1}
    \end{figure}

\begin{figure}[h]
        \centering
        \begin{subfigure}[b]{0.23\textwidth}
             \centering
             \includegraphics[width=\textwidth]{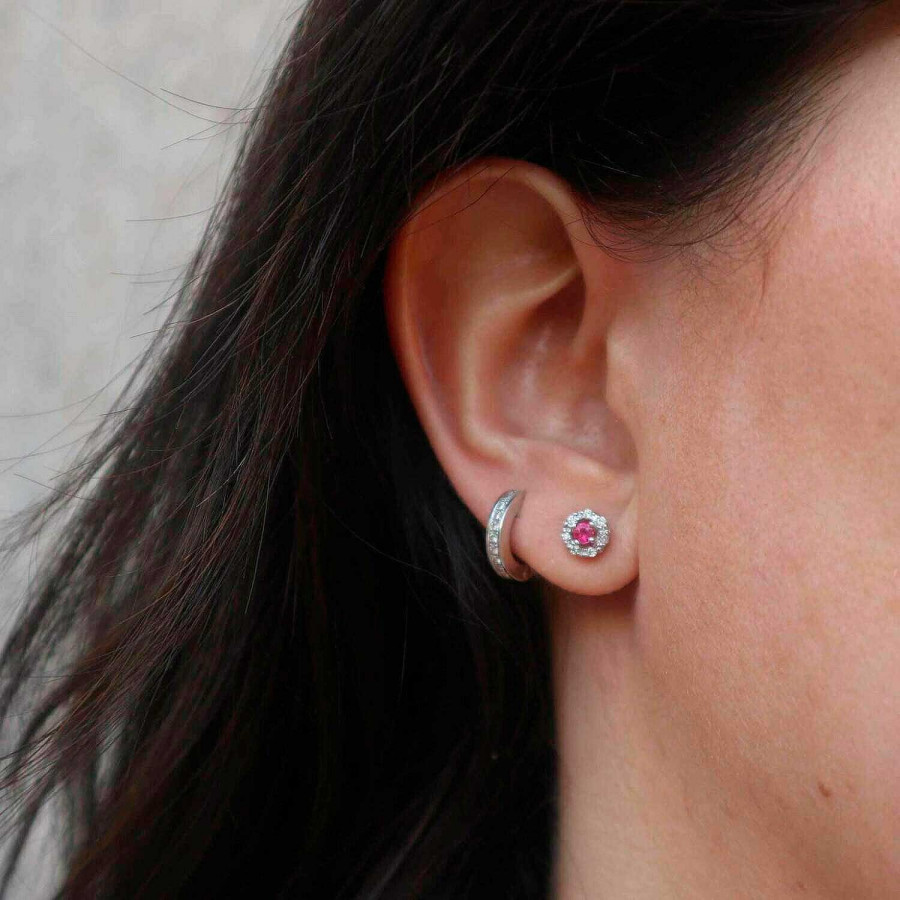}
             \caption{Badly descripted accessory ``Orion ruby earring''.}
             \label{subfig:bad3}
        \end{subfigure}
        \begin{subfigure}[b]{0.23\textwidth}
             \centering
             \includegraphics[width=\textwidth]{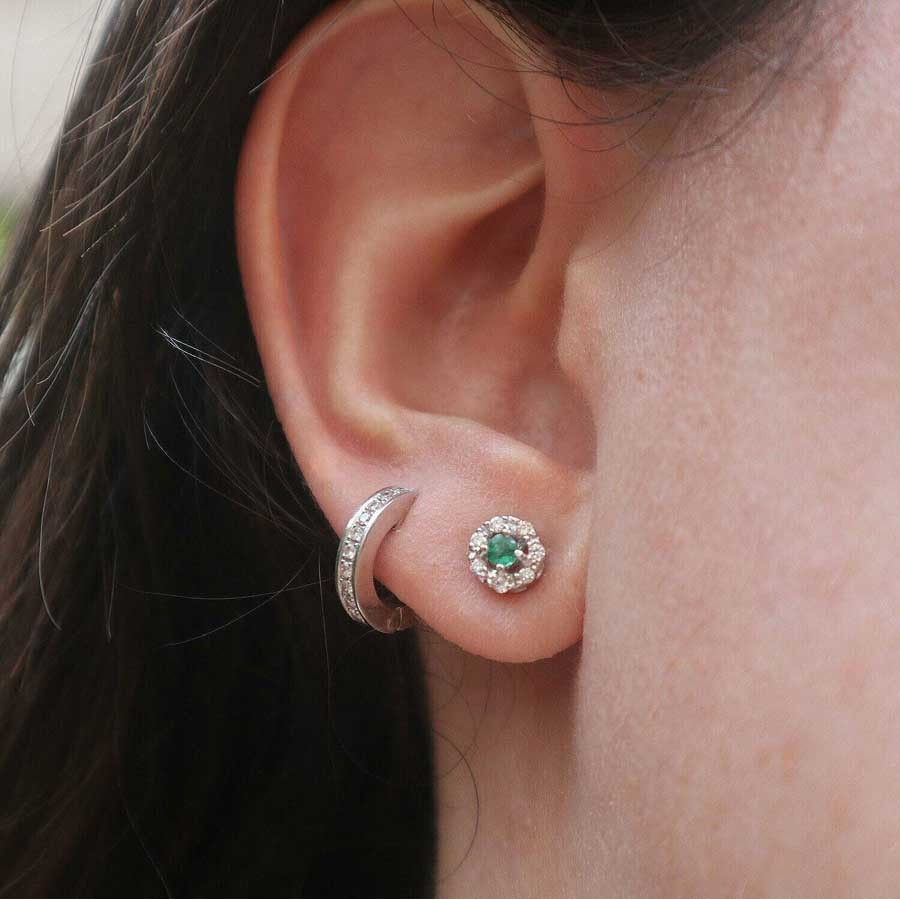}
             \caption{Descripted as ``Orion emerald earring''.}
             \label{subfig:bad4}
        \end{subfigure}
        \caption{Accessories badly described by the best model. Different materials.}
        \label{fig:mal_2}
    \end{figure}

Finally, a web page interface has been also created using the best image captioning developed models. This interface allows the user to upload an accessory image to generate a description about it. The web page interface is shown in Figure \ref{fig:web}. The description obtained can be generated using 3 different complexity levels:

\begin{itemize}
    \item Basic description: describes only which type of accessory is shown in the picture (bracelets, rings, earrings, etc.).
    \item Normal description: describes the type of accessory and its colors and materials.
    \item Complete description: describes the type of accessory, its colors and materials, and also identifies its specific model.
\end{itemize}

\begin{figure*}[h]
    \centering
    \includegraphics[width=0.8\textwidth]{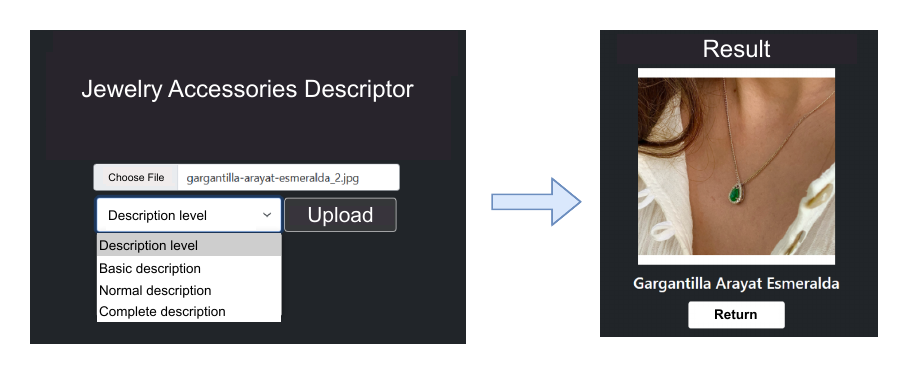}
    \caption{Web page interface for accessory image captioning. Description for a necklace input image.}
    \label{fig:web}
\end{figure*}

Presently, the model has been handed over to jewellers for testing purposes. However, it is not yet deemed ready for practical implementation. Nevertheless, once the dataset is enhanced, it is anticipated that more robust models will emerge, enabling a multitude of applications. These applications include automated generation of accessory descriptions for websites and providing detailed descriptions of jewelry to potential customers via a mobile app, aiding their purchasing decisions.

\section{Acknowledgements}

José Manuel Alcalde Llergo is a PhD student enrolled in the National PhD in Artificial Intelligence, XXXVIII cycle, course on Health and life sciences, organized by Università Campus Bio-Medico di Roma. This PhD is also been performed in cotutoring with the University of Córdoba.

\bibliographystyle{ieeetr}
\bibliography{bibliography}\clearpage

\end{document}